\def\year{2020}\relax
\documentclass[letterpaper]{article}
\usepackage{aaai20}
\usepackage{times}
\usepackage{helvet}
\usepackage{courier}
\usepackage{url}
\usepackage{graphicx}
\usepackage[utf8]{inputenc} 
\usepackage{url}    
\usepackage{booktabs}       
\usepackage{amsfonts}       
\usepackage{nicefrac}       
\usepackage{microtype}      
\usepackage{algorithm, algorithmic}
\usepackage{bm}
\usepackage{enumerate}
\usepackage{paralist}

\usepackage{amsmath,amssymb}
\usepackage{amsthm}

\newtheorem{theorem}{Theorem}[section]
\newtheorem{corollary}{Corollary}[theorem]
\newtheorem{lemma}[theorem]{Lemma}
\theoremstyle{definition}
\newtheorem{definition}{Definition}[section]
\DeclareMathOperator*{\argmax}{arg\,max}
\DeclareMathOperator*{\argmin}{arg\,min}

\newcommand{\cen}{\rho}
\newcommand{\probname}{Hierarchical-Revenue }
\newcommand{\probnameshort}{Hierarchical-Revenue}
\newcommand{\hc}{hierarchical clustering }
\newcommand{\ckmm}{\textsc{ckmm }}

\newcommand{\citet}[1]{\citeauthor{#1} \shortcite{#1}}
\newcommand{\citep}{\cite}

\newenvironment{proofof}[1]{\smallskip\noindent{\bf Proof of #1}}%
        {\hspace*{\fill}$\qed$\par}

\title{An Objective for Hierarchical Clustering in Euclidean Space and its Connection to Bisecting K-means\thanks{Y. Wang and B. Moseley were supported in part by a NSF Grants CCF-1830711, CCF-1733873, CCF-1733873 and CCF-1845146, a Google Research Award, a Bosch junior faculty chair  and an Infor faculty award.}}

\author{
Yuyan Wang\textsuperscript{\rm 1} \and
Benjamin Moseley\textsuperscript{\rm 1}\\
\textsuperscript{\rm 1} Tepper School of Business, Carnegie Mellon University, Pittsburgh, PA \\
$\{$yuyanw,moseleyb$\}$@andrew.cmu.edu 
}

\author{\Large \textbf{Yuyan Wang\textsuperscript{\rm 1}, Benjamin Moseley\textsuperscript{\rm 1}}\\ 
\textsuperscript{\rm 1}Tepper School of Business, Carnegie Mellon University\\ 
5000 Forbes Avenue\\
Pittsburgh, Pennsylvania 15213\\
$\{$yuyanw, moseleyb$\}$@andrew.cmu.edu  
}

\begin{document}
%
\maketitle
\begin{abstract}
\begin{quote}
This paper explores hierarchical clustering in the case where pairs of points have dissimilarity scores (e.g. distances) as a part of the input.  The recently introduced objective for points with dissimilarity scores results in \emph{every tree} being a $\frac{1}{2}$ approximation if the distances form a metric. This shows the objective does not make a significant distinction between a good and poor hierarchical clustering in metric spaces. 
  
Motivated by this, the paper develops a new global objective for hierarchical clustering in Euclidean space.  The objective captures the criterion that has motivated the use of divisive clustering algorithms: that when a split happens, points in the same cluster should be more similar than points in different clusters. Moreover, this objective gives reasonable results on ground-truth inputs for hierarchical clustering.

The paper builds a theoretical connection between this objective and the bisecting $k$-means algorithm. This paper proves that the optimal $2$-means solution results in a constant approximation for the objective. This is the first paper to show the bisecting $k$-means algorithm optimizes a natural global objective over the entire tree. 

\end{quote}
\end{abstract}
\section{Introduction}
\label{intro}


In hierarchical clustering, the input is a set of points, with a score that represents the pairwise similarity or dissimilarity of the points. The goal is to output a tree, often binary, whose leaves represent data points, and internal nodes represent clusters. Each internal node is a cluster of the leaves in the subtree rooted at it.  When a node gets closer towards the leaves, the cluster it represents should become more refined, and the points in this cluster should become more similar.   The nodes of the same level in this tree represent a partition of the given data set into clusters.  Note that each data point (leaf) belongs to many clusters, one for each ancestor.

\begin{figure}[H]
\centering 
\includegraphics[height=0.14\textwidth]{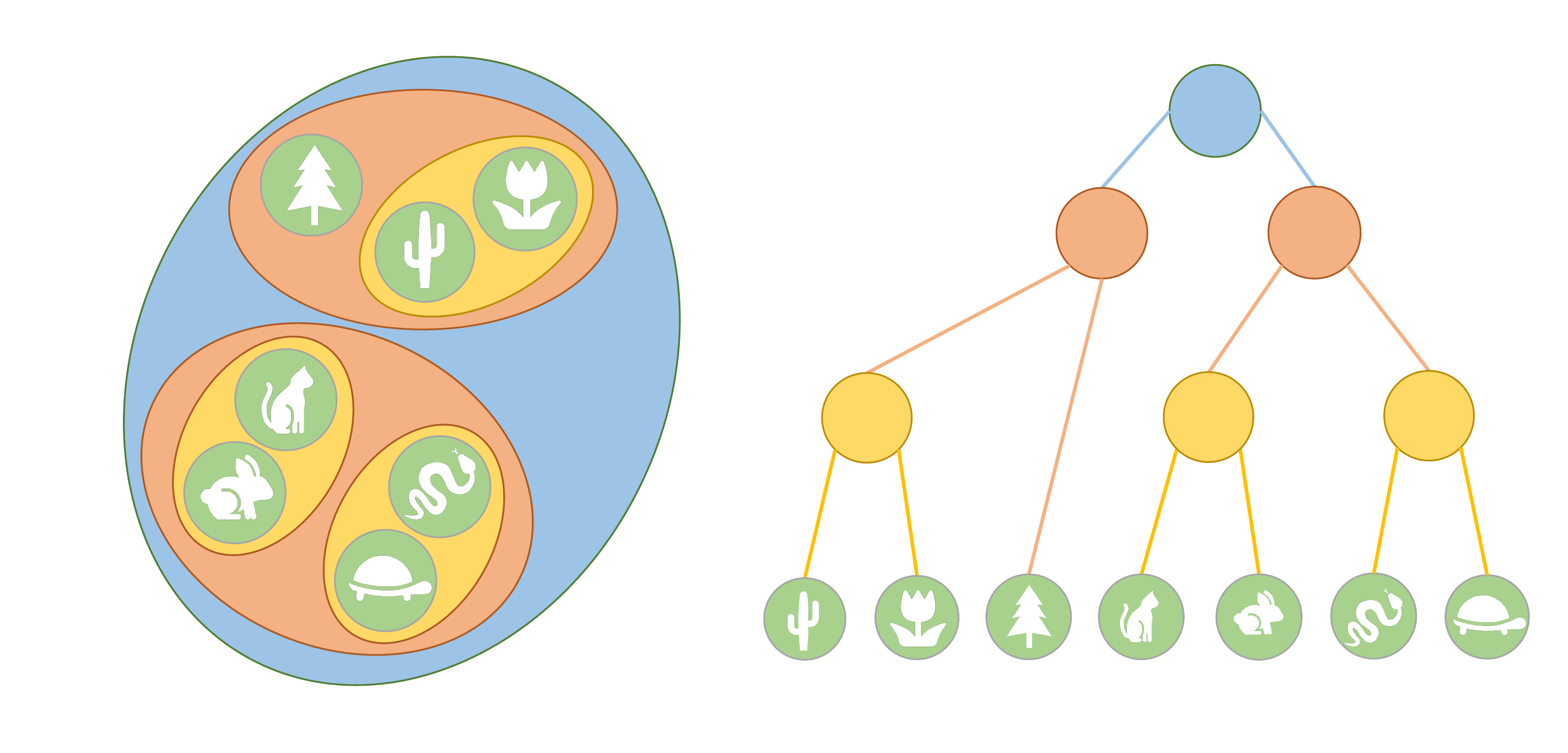} 
\caption{A hierarchical clustering tree.   The green leaves are the input data points. Internal nodes represent a cluster of the leaves in their subtree. }
\label{tree} 
\end{figure}

Figure \ref{tree} shows a sample hierarchical clustering tree for biological species. All the nodes in the tree on the right are painted in the same colors with the clusters they represent in the picture on the left.

The mainstream algorithms used to do hierarchical clustering can be classified roughly into two categories: agglomerative and divisive. \textbf{Agglomerative algorithms} initialize every point to be in their own individual cluster. They iteratively pick the two clusters that are the most similar to each other to merge into a bigger cluster. Meanwhile, the algorithms create a parent in the hierarchical tree produced that is connected to the two nodes corresponding to the two clusters before merging. The merging terminates when only one cluster remains.

 It is necessary for an agglomerative algorithm to quantify the similarity between clusters, which can be defined in several ways. For example, average-linkage is a popular agglomerative algorithm, which measures the similarity of two clusters by calculating the average pairwise inter-cluster similarity score. \citet{steinbach2000comparison}, \citet{DBLP:journals/cj/Murtagh83}, \citet{DBLP:journals/widm/MurtaghC12} and \citet{DBLP:journals/datamine/ZhaoKF05} discussed about common agglomerative algorithms and compared the performance of different agglomerative algorithms in a variety of backgrounds. \citet{DBLP:journals/jmlr/AckermanB16} identified properties of trees produced by linkage-based agglomerative algorithms.

\textbf{Divisive algorithms} initialize the whole point set as one single cluster, and create a root node corresponding to this cluster in the hierarchical tree. They iteratively \textbf{split} a cluster into smaller clusters. Then the algorithms create  nodes representing the separated clusters in the tree and make them the children of the parent node. During the split, we want the points in different clusters to be less similar than the points in the same cluster. Again, the notion of similarity is open to different interpretations. A divisive algorithm terminates when every point is in its own individual cluster. The divisive algorithms that split a set into two subsets at each iteration are called bisecting algorithms. 


Data scientists regularly use the bisecting $k$-mean algorithms at each level\footnote{The word ``bisecting" refers to the case when $k=2$.}.  This is used when the distances between the data points are used as  dissimilarity scores.  See \citet{steinbach2000comparison} and \citet{DBLP:journals/widm/MurtaghC12} for more information on divisive algorithms. \citet{DBLP:journals/jmlr/AckermanB16} proved that popular divisive algorithms can produce clusterings different from linkage-based agglomerative algorithms.  Naturally, they may be optimizing fundamentally different criteria.  

\medskip
\noindent \textbf{Objective Functions:} There has been a recent interest in identifying a global objective for hierarchical clustering. \citet{DBLP:conf/stoc/Dasgupta16} developed a cost function objective for data sets with similarity scores between points. For a given data set $V$ with $n$ points $\{1,2,3,...,n\}$, let $w_{ij}$ be the similarity score between points $i$ and $j$. In a tree $T$, let $T[i \lor j]$ denote the subtree rooted at the least common ancestor of $i$ and $j$, and $|\verb+leaves+(T[i \lor j])|$ denote the number of leaves of $T[i \lor j]$. The cost objective function objective introduced in \citet{DBLP:conf/stoc/Dasgupta16} is defined as: $\min_{T} cost_T(V)=\sum_{1 \leq i < j \leq n}w_{ij}|\verb+leaves+(T[i \lor j])|$.

Since every $w_{ij}$ is multiplied with the number of leaves of the smallest tree containing both $i$ and $j$, the points that are more similar  (bigger $w_{ij}$'s) are encouraged to have $T[i \lor j]$ with fewer leaves. In other words, the objective function is encouraging points which are more similar to each other to be split at lower levels of the tree where there are fewer leaves at the least common ancestor.

The work of \citet{DBLP:conf/stoc/Dasgupta16} has initiated an exciting line of study \cite{RoyP17,cohen2019hierarchical,pmlr-v89-charikar19a,ChatziafratisNC18,abs-1811-00928}. \citet{cohen2019hierarchical} generalized the results in \citet{DBLP:conf/stoc/Dasgupta16} into a class of cost functions that possess properties desirable of a valid objective function.   They showed that the average-linkage algorithm is a $\frac{2}{3}$-approximation for an objective based on the \citet{DBLP:conf/stoc/Dasgupta16} objective.\footnote{Throughout this paper we use $c > 1$ for approximations on minimization problems and $c < 1$ for maximization.} This objective modifies the Dasgupta objective to handle dissimilarity scores. Let $d(i,j)$ be the distance between $i$ and $j$.  The objective is $\max_T \sum_{1 \leq i < j \leq n}d(i,j)|\verb+leaves+(T[i \lor j])|$. The motivation is similar to the Dasgupta objective, except now the similarity score $w_{ij}$ is swapped to a dissimilarity score $d(i,j)$ and the problem is changed to a maximization problem. Contemporaneously, \citet{DBLP:conf/nips/MoseleyW17} designed a revenue objective function based on  \citet{DBLP:conf/stoc/Dasgupta16} and showed the average-linkage algorithm is a constant approximation for the objective. \citet{pmlr-v89-charikar19a} showed an improved analysis of average-linkage  for Euclidean data. Together \citet{cohen2019hierarchical}, \citet{pmlr-v89-charikar19a}  and \citet{DBLP:conf/nips/MoseleyW17} have established a relationship between a practically popular algorithm and global objectives. This gives a foundational understanding of the average-linkage algorithm.

\noindent \textbf{Euclidean Data:} This paper is interested in data embedded in Euclidean space where the $\ell_2$ distance between points represents their dissimilarity.    There is currently one global objective that  has been proposed for data with dissimilarity scores.  This is the objective of \citet{cohen2019hierarchical} described above, an extension of the Dasgupta objective \cite{DBLP:conf/stoc/Dasgupta16}. Throughout this paper, we refer to the objective in \citet{cohen2019hierarchical} as \ckmm objective. This paper shows in Section \ref{sec:everytree}  that \emph{every} tree is a $\frac{1}{2}$-approximation for the \ckmm objective if the data in a metric space. Previously, it was known that all trees gave a constant approximation \cite{cohen2019hierarchical}.

In a common case where data is in Euclidean space, one type of metric, the objective does not make a large differentiation between different clusterings.  In practice, it is clear that some trees are more desirable than others. It is an interesting question to find an objective that makes a stronger distinction between different clusterings. This is the target question this paper addresses.

\noindent \textbf{Divisive Algorithms:}  While great strides have been made on the foundations of hierarchical clustering, it remains an open question to explain what popular divisive algorithms optimize.   In particular, the popular bisecting $k$-means algorithm has been proven to be at least a factor $O(\sqrt{n})$ far from optimal for the objectives given in \citet{DBLP:conf/nips/MoseleyW17} and \citet{DBLP:conf/stoc/Dasgupta16}. This can be viewed as these algorithms being \emph{extremely bad} for these objectives in the worst case. This contrasts with  the performance of average-linkage for known objectives.  Perhaps, this highlights that bisecting $k$-means and other divisive algorithms optimize something fundamentally different than average-linkage and general linkage based algorithms.   It remains to discover a global objective that helps characterize  the optimization criteria of divisive algorithms, another target of this paper. 

\medskip
\noindent \textbf{Results:} This paper introduces a new revenue maximization objective for hierarchical clustering on a point set in Euclidean space.  The objective is designed to capture the main criterion that motivates the use of divisive algorithms: when data is split at a level of the tree, the data in each sub-cluster should be closer to each other than data points in different clusters. 

Each node in the tree corresponds to a \emph{split} that generates revenue.  The objective specifies that the global revenue of the tree is the summation of the revenue at each node.    The split revenue captures the quality of the split.

\begin{figure}[H] \label{fig:intra_and_inter}
\centering 
\includegraphics[height=2cm]{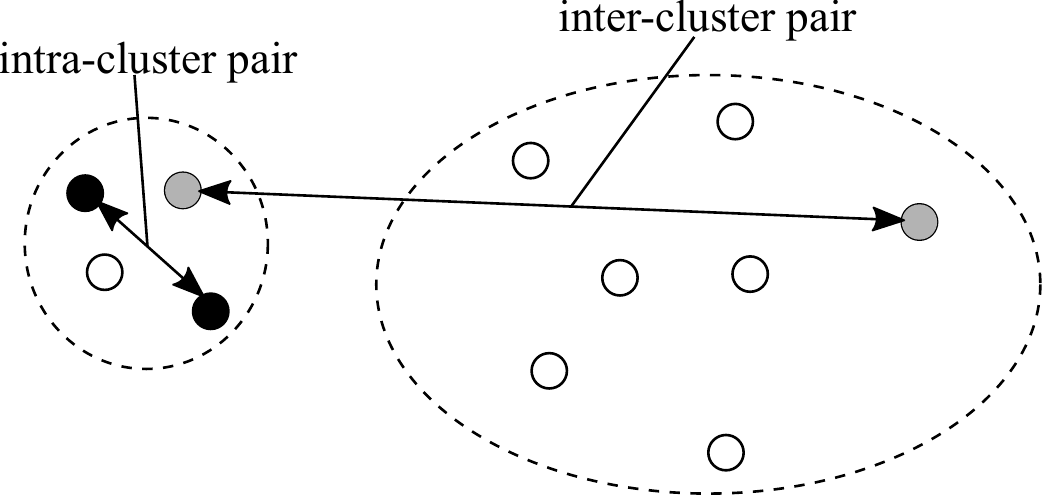} 
\caption{Intra- and inter- cluster distance of two clusters. The black pair is an example of intra-cluster pairs, and the grey pair is an example of inter-cluster pairs. }
\end{figure}

\begin{itemize}
\item \textbf{Guiding Principle:} The new objective function enforces that a split is good if the \emph{inter-cluster distances are big compared to intra-cluster distances \footnote{Here ``inter-cluster distances" refers to that between points in different clusters, while ``intra-cluster distances" refers to that between points in the same cluster.}}, as is indicated in Figure \ref{fig:intra_and_inter}. This is the main motivation behind a generic divisive algorithm.  Of course, the global tree structure influences the possible revenue at an individual split.
\end{itemize}

We show several interesting properties of this new objective.  

\begin{itemize}
\item For problem instances corresponding a ground-truth  as introduced in \citet{cohen2019hierarchical}, this objective gives desirable optimal solutions. In particular, \citet{cohen2019hierarchical} introduced a large class of  instances that have a natural corresponding \hc that should be optimal.  We prove that these trees are \emph{optimal} for the new objective function we propose on such instances. We note that these instances generalize instances given in \citet{DBLP:conf/stoc/Dasgupta16} that were used to motivate a hierarchical clustering objective. 
\item The bisecting $k$-means algorithm is a constant approximation for the objective. This establishes that the objective is closely related to the bisecting $k$-means algorithm and aids in understanding the underlying structure of solutions the algorithm produces. \emph{This is the first global objective that this algorithm is known to provably optimize.}
\item The objective is trivially modular over the splits, like the objectives of  \citet{cohen2019hierarchical}, \citet{DBLP:conf/nips/MoseleyW17} and \citet{DBLP:conf/stoc/Dasgupta16}.
\item In the context of metric spaces, this objective has different properties compared to some proposed objectives. It is known that the Random algorithm\footnote{See Section~\ref{sec:random} for a  formal description of the algorithm.}, which partitions data uniformly at random at each node, is a constant approximation for the \ckmm objective with dissimilarity scores that need not form a metric. Further, it is a constant approximation for the  \citet{DBLP:conf/nips/MoseleyW17} objective with similarity scores. For these two  objectives, Random is a $\frac{2}{3}$ and $\frac{1}{3}$ approximation, respectively. The Random  algorithm can produce undesirable hierarchical clusterings and it is counterintuitive that it is a constant approximation for these objectives.  This paper shows that Random results in an $O(\frac{1}{n^\epsilon})$-approximation for the proposed objective for a constant $\epsilon >0$.  Therefore, the Random algorithm provably performs poorly for the new objective.  This can be seen as a strength of the new objective over those proposed. 
\end{itemize}
We further show the following about other objectives in metric space. These show that some other objectives do not make a large differentiation between trees in metric space, even if the trees correspond to a poor clustering.  Our objective does and this can be seen as an advantage of the new objective.
\begin{itemize}
\item As mentioned, we show that every tree is a $\frac{1}{2}$-approximation for \ckmm objective  when points have dissimilarity scores that form a metric. 
\item We show that  every tree is a $2$-approximation for the Dasgupta objective  \cite{DBLP:conf/stoc/Dasgupta16} for similarity scores that satisfy the triangle inequality.  We include this result to provide insight into this objective.  However we note that this is less surprising than the similar result on the \ckmm objective since some natural similarity score instances do not satisfy the triangle inequality.

\end{itemize}

We investigate empirically the performance of three popular algorithms used in practice and Random algorithm for the new objective. As is suggested by theory, the proposed objective moderately favors bisecting k-means over two agglomerative algorithms, while magnifying the gap between the performance of Random and the other three algorithms.

\noindent \textbf{Other Related Work:} Other work centers around when bisecting algorithms work well.   The work of \citet{DasguptaL05,Plaxton06} show the remarkable result that hierarchical trees exists such that each level of the tree optimizes the corresponding $k$-clustering objective.  These algorithms are complex and are mostly of theoretical interest.   \citet{BalcanBV08} showed that partitioned clusterings can be uncovered from using hierarchical clustering methods under stability conditions. The work of \citet{AwasthiBS12,BalcanBG13,CarlssonM10} and pointers therein study stability conditions of clustering.

The work of \citet{CharikarC17} and \citet{RoyP17} were the first to give improved bounds on the objectives of Dasgupta. Currently, the best known approximations for both the objective of \citet{DBLP:conf/stoc/Dasgupta16} and \ckmm \cite{cohen2019hierarchical} were shown in \citet{CharikarCN19}.  This work gave a $\frac{1}{3} +\epsilon$ and $\frac{2}{3} + \delta$ approximations for some small constants $\epsilon$ and $\delta$, respectfully. Further, they shown that average-linkage is no better than a $\frac{1}{3}$ and $\frac{2}{3}$ approximation for the objectives respectively.  Thus, new algorithms were required to improve the approximation ratio. If there data is in Euclidean space, then \citet{pmlr-v89-charikar19a} gave improve approximation ratios.  

\section{Preliminaries}
\label{sec:prelim}
In this section, we give a formal mathematical definition for the hierarchical clustering problem. Then the objective function is given. 

\noindent \textbf{Problem Input}: In the hierarchical clustering problem, the input is a set $V$ of data points. There is a distance between each pair of points $i$ and $j$ denoting their dissimilarity. In this paper, the data points are assumed to be located in Euclidean spaces, one particular type of metric space. For each pair of points $(i,j)$, the $\ell_2$ distance, denoted as $d(i,j)$, is used as their dissimilarity score. The $\ell_2$ distance is known to satisfy the following properties:
\begin{compactenum}[(1)]
\item Convexity. The distances satisfy Jensen's inequality: for any points $i,j,k$, $d(\lambda \cdot i + (1-\lambda)\cdot j,k) \leq \lambda d(i,k) + (1-\lambda) d(j,k)$, where $\lambda \in [0,1]$.
\item Triangle inequality. For any points $i,j,k$, $d(i,k) \leq d(i,j)+d(j,k)$.
\end{compactenum}



\noindent \textbf{$k$-means Objective}: The definition of the $k$-means objective is the following. Given a point set $S$, a $k$-means clustering partitions $S$ into $k$ sets $S_1, S_2, \ldots S_k$. The $k$-means objective calculates the summation over the squared norm of the distance between a point to the \textbf{centroid} of the set it belongs to: $\sum_{j=1}^{k} \sum_{u \in S_j} d^2(u,\cen(S_j))$. Here $\cen(S_j)$ denotes the centroid of $S_j$. In Euclidean space, $\cen(S_j)$ satisfies: $\cen(S_j)=\frac{\sum_{u \in S_j} u}{|S_j|}=\argmin_{p}\sum_{u \in S_j}d^2(u,p)$.



Let $\Delta_k(S)$ denote the optimal $k$-means objective function value for the point set $S$, where $k$ is the number of clusters.  We will be particularly interested in $\Delta_2(S)$, the $2$-means objective. 

Fix a hierarchical clustering tree $T$ on a set $V$. Consider a node of the tree and let $S\subseteq V$ be the subset of input data that is input to the current split.  These will eventually be the leaves of the subtree induced by this node. We use $S \rightarrow (S_1, S_2)$ to denote a split in the tree where a set $S$ is separated into two non-empty subsets, $S_1$ and $S_2$.  These sets correspond to the input of the two child nodes. We let $S \rightarrow (S_1, S_2) \in T$ denote that this split exists in $T$.     

Any split $S \rightarrow (S_1, S_2)$ where $S_1$ and $S_2$ are a partition of $S$ is a valid $2$-means solution for the point set $S$. Since $\Delta_2(S)$ denotes the optimal objective function value, $\Delta_{2}(S)  \leq \Delta_{1}(S_1) + \Delta_{1}(S_2)$ by definition of the $2$-means objective. In particular, if $S \rightarrow (S_1, S_2)$ is the optimal $2$-means solution, we have $\Delta_{2}(S) = \Delta_{1}(S_1) + \Delta_{1}(S_2)$.




\section{\probname: Comparing Inter vs. Intra Cluster Distance }

This section defines the new objective function. We call the problem of optimizing this objective the \textbf{\probname problem}.  

\noindent \textbf{Defining the Revenue for a Pair:}  Consider a node in a hierarchical clustering whose input is $S$ and this set is split into $S_1$ and $S_2$.  A good tree ensures that the pairs of points in $i,i' \in S_1$ (resp. $S_2$) are more similar that pairs $i \in S_1$ and $j \in S_2$ (i.e. $d(i,j) \geq d(i,i')$).  This ensures the points corresponding to the cluster at a node in the tree become more similar at lower levels of the tree.  In the following, we say $i$ and $j$ are \textbf{split} the first time they no longer belong to the same cluster.   

Every pair $i$ and $j$ will be eventually split in the tree and a hierarchical clustering objective should ensure they are split at the appropriate place in the tree.  Further, an objective should  optimize over all pairs uniformly to determine the splits. 

Guided by these principles, we develop the objective as follows.  We begin by allowing every pair $i$ and $j$ to generate one unit of revenue.  This revenue can always be obtained for a fixed pair, but not necessarily for all pairs simultaneously.  This unit of revenue is obtained when the pair is split at an appropriate position in the tree.  Less revenue (or even 0) will be obtained when the pair is separated at a poor position. \emph{This is the key to determine the quality of a split.}

Say that $S \rightarrow (S_1, S_2)$ is the split at some node in the tree and $i \in S_1$ and $j \in S_2$ are split. As discussed above, points in $S_1$ (respectively $S_2$) should be more similar to each other than $i$ and $j$.  To measure the similarity of $i$ to other points in $S_1$ we use $d(i,\cen(S_1))$, the distances of $i$ to the centroid of $S_1$.  Similarly, we use $d(j, \cen(S_2))$ to measure the distance of $j$ to points in $S_2$.  The distance of a point to the centroid of a set measures the  distance to the average point in the set.  Thus, we would like $d(i,j)$ to be larger than both  $d(i,\cen(S_1))$ and $d(j, \cen(S_2))$ for it to make sense to split $i$ and $j$.  That is, $i$ and $j$ should become more similar to their respective sets after the split than they are to each other.

 Formally, define the revenue for a pair of points as follows. Let $\delta_{S_1,S_2}(i,j) = \max\{d(i,\cen(S_1)), d(j, \cen(S_2)) \}$ be the maximum distance of $i$ and $j$ to their respective centroids.  We would like $\delta_{S_1,S_2}(i,j)$ to be smaller than $d(i,j)$ and therefore $i$ and $j$ generate a unit of revenue when this is the case.  When $\delta_{S_1,S_2}(i,j) \leq d(i,j)$ we assume the revenue decays linearly.  That is, the revenue is  $\frac{d(i,j)}{\delta_{S_1,S_2}(i,j)}$. 
 
 Putting the above together, define the revenue for splitting $i$ and $j$ as $rev(i,j) = \min\{\frac{d(i,j)}{\delta_{S_1,S_2}(i,j)},1\}$\footnote{We assume dividing by $0$ gives revenue $1$.}. This is the \textbf{revenue} $i$ and $j$ generates.  Notice that a revenue of a unit can always be obtained since we can let $j$ be the unique last point split from $i$. However, a good hierarchical splitting structure is needed to get good revenue for many pairs of points.
 
 \medskip
\noindent \textbf{The Global Objective:} The global objective is defined as follows.  We note that while the revenue is summed over each split in the tree, obtaining a large amount of revenue at a split hinges on a good global tree structure. 

 \begin{definition}[\probnameshort]
\label{revenue_functions_means}
For a data set $V$ and a given hierarchical clustering tree $T$, define the hierarchical tree revenue function as follows. Let $rev(S_1, S_2) = \sum_{i \in S_1} \sum_{j \in S_2} rev(i,j) =\sum_{i \in S_1} \sum_{j \in S_2}  \min\{\frac{d(i,j)}{\delta_{S_1,S_2}(i,j)},1\}$ be the revenue over all pairs of points split across $S_1$ and $S_2$. The aggregate revenue is   $rev_T(V) =\sum_{\{i,j\} \subseteq V}  rev(i,j) = \sum_{S \rightarrow(S_1,S_2) \in T}rev(S_1,S_2)$, and it should be maximized over all trees.
\end{definition}
 
As is shown in Definition \ref{revenue_functions_means}, there are two ways of computing $rev_T(V)$. One is to sum up the  revenue over the pairs, while the other is to sum up the revenue over the splits. Both methods lead to the same value. The second form allows us to judge whether a  split at some internal node of the tree is good or not compared to the number of pairs it separated.



\section{Ground-truth Inputs}
The work of \citet{cohen2019hierarchical} gave a characterization of desirable \hc objectives. The idea is to give a class of instances that naturally correspond to a specific  \hc tree. These trees should be optimal solutions for a good \hc objective.


 In particular, \citet{cohen2019hierarchical}  defined input instances that correspond to \emph{ultrametrics}. Such inputs will be referred to as \emph{ground-truth} inputs.  For such an input, they define \emph{generating trees}, which should be optimal for the \hc objective to be valid.  Intuitively, in an ultrametric either it is clear what the split should be at each point in the tree or all splits are equivalent.\footnote{If there is a natural split then the points can be divided into two groups $A$ and $B$ such that inter-group distances are larger than intra-group distances.  If all splits are equivalent then pairwise the points are all the same distance.} The resulting tree is a generating tree. 

We prove a generating tree is an optimal solution for our objective, if the input in Euclidean space is ground-truth. 

\subsection{Definition of Ground-Truth Inputs}
We cite the following definitions from \citet{cohen2019hierarchical}. 
\begin{definition} \label{def:ultrametric}
A metric space $(X,d)$ is an ultrametric if for every $x,y,z \in X$, $d(x,y) \leq \max \{d(x,z),d(y,z)\}$.
\end{definition}

Intuitively,  the definition of ultrametric implies that any three points $u,v,w$ form an isosceles triangle, whose equal sides are at least as large as the other side. \citet{cohen2019hierarchical} then defined an instance generated from ultrametric, which is treated as ground-truth input for hierarchical clustering. 

\begin{definition} \label{def:generate_ultrametric}
An input instance on a set of points $V$ with pairwise distance function $d$ is \emph{generated from an ultrametric} if the distances function $d$ corresponds to a ultrametric.
\end{definition}

Following \citet{cohen2019hierarchical}, we define \emph{generating trees}, which are considered the most well-behaving \hc trees for a ground-truth input.
\begin{definition} \label{def:generating_tree}
If the instance $V$ is generated by ultrametric, a binary tree $T$ is a generating tree for $G$ if it satisfies the following properties:
\begin{enumerate}
\item It has $|V|$ leaves and $|V| - 1$ internal nodes. Let $L$ denote its leaves and each point in $L$ corresponds to a unique point in $V$. Let $\mathcal{N}$ denote its internal nodes, corresponding to clusters of the leaves of the subtree rooted at the node.
\item There exists a weight function $W: \mathcal{N} \mapsto \mathbb{R}_+$. For $N_1,N_2 \in \mathcal{N}$, if $N_1$ is on the path from $N_2$ to the root, $W(N_1) \geq W(N_2)$. For every $x,y \in V$, $d (x,y)  = W(LCA_T(x,y))$, where $LCA_T(x,y)$ denotes the Least Common Ancestor of leaves corresponding to $x$ and $y$ in $T$.
\end{enumerate}
\end{definition}

\citet{cohen2019hierarchical} proposed that for a ground-truth input graph as defined in Definition~\ref{def:generate_ultrametric}, if there exists any corresponding generating tree $T$, it is considered one of the best solutions among all the solutions, and thus should be one of the optimal solutions for the \hc objective function used.

We give some  intuition for why a generating tree is considered the best tree on such inputs. The value $W(LCA_T(x, y))$ can be interpreted as the  distances of edges cut in split at the LCA of $x$ and $y$. All points separated in a split have equal pairwise distance, the maximum pairwise distances in the current point set. Naturally, the higher up this LCA is, the larger the distance should be. For each ground-truth input graph, there is always a generating tree $T$, which separates the farthest pairwise points in every split.

\subsection{Optimality of Generating Trees}
Now we prove that given an input that is generated from an  ultrametric, every generating tree is optimal for \probname function introduced in this paper. In particular, every pair of points will get full revenue.

\begin{lemma} \label{lem:generating_tree_cuts_biggest_distances}
A binary tree $T$, with $|V|$ leaves corresponding to the points in $V$ and $|V|-1$ internal nodes, is a generating tree for an instance $V$ generated from a ultrametric  if and only if it satisfies the following property:
\begin{itemize}
    \item For every split $A \cup B \rightarrow (A,B)$ from top to bottom, $\forall{i \in A, j \in B}, d(i,j) = \max_{x \in A, y \in B }d(x,y)$.
\end{itemize}
\end{lemma}

Every ground-truth input has at least one generating tree, as stated in the following theorem.

\begin{theorem} \label{thm:always_generating_tree}
For every instance generated from some ultrametric, there is always a generating tree $T$ as defined in Definition~\ref{def:generating_tree}.
\end{theorem}

Using Lemma~\ref{lem:generating_tree_cuts_biggest_distances}, the optimality of $T$ is proved by arguing every split gives a revenue of $1$ for every pair of points it separates.

\begin{theorem} \label{thm:generating_tree_optimal}
A generating tree $T$ for an instance generated $V$ from an ultrametric  is optimal for the \probname objective. 
\end{theorem}

\section{Bisecting $k$-means Approximates the Revenue Objective}
\label{sec:2_means_approx}
This section shows that the bisecting $k$-means algorithm is a constant approximation for the proposed objective.   This establishes a foundational connection between a natural objective function and the bisecting $k$-means algorithm.  This is the first analysis showing that bisecting $k$-means optimizes a global objective function.    This helps explain the structure of the solutions produced by the algorithm.

The goal of this section is to show the following theorem. 
\begin{theorem}\label{thm:mainbisect}
Fix any input set $V$ and let $T$ be the tree created by the bisecting $k$-means algorithm. The tree $T$ is a constant approximation for the \probname objective.
\end{theorem}

The analysis is based on analyzing each split performed by bisecting $k$-means individually.  The following lemma shows that if every split in a hierarchical clustering tree is good for the objective function proposed, then the whole tree is also good. By ``good" we mean that the split gains a revenue which is at least some constant factor times the number of pairs separated.  This lemma follows immediately by definition of the objective. 

\begin{lemma}\label{analysis_structure} 
A hierarchical clustering tree $T$ is a $\gamma$-approximation for the \probname problem if it satisfies the following condition: $\forall S \rightarrow (S_1, S_2) \in T$, $rev(S_1, S_2) \geq \gamma |S_1||S_2|$ holds for some constant $\gamma>0$.
\end{lemma}

The above lemma allows us to focus on a single iteration of the bisecting $k$-means algorithm. Suppose at some iteration, a cluster $A \cup B$ is split into $A$ and $B$. We give the following definition of a \emph{high-revenue point}. A point $u$ in $A$ is a high-revenue point if for most of the points in $B$, it gains acceptable amount of revenue. 

\begin{definition}\label{def:good_set}
Given a split $A \cup B$ and a point $u \in A$, the \textbf{high-revenue set} for $u$ for set $B$ is:
$HR_B(u)=\{v \in B:rev(u,v) \geq \frac{1}{10}\}.$
Define the \textbf{low-revenue set} for $u \in A$ to be defined as:
$LR_B(u)=\{v \in B:rev(u,v) < \frac{1}{10}\}=B \setminus HR_B(u)$.
\end{definition}
\begin{definition}\label{def:good_point}
Given a split $A \cup B$, a point $u \in A$ is a \textbf{high-revenue point} if $|HR_B(u)| \geq \frac{1}{2}|B|$. Otherwise, it is called a low-revenue point.
\end{definition}

With the definition of high-revenue points in place, the next lemma claims that given split $A \cup B \rightarrow (A,B)$ created by the optimal $2$-means algorithm, if $|A| \geq |B|$, at least half of $A$ are high-revenue  points.  This is the main technical lemma.  This combined with Lemma~\ref{analysis_structure} implies Theorem~\ref{thm:mainbisect}.

\begin{lemma}\label{lem:main_theorem}
Let $A$ and $B$ be the optimal $2$-means solution for the point set $A \cup B$. Without loss of generality, suppose $|A| \geq |B|$. Then, at least $\frac{4}{7}|A|$ points in $A$ are high-revenue. This gives a lower bound of at least $\frac{1}{35}|A||B|$ revenue in total for splitting $A$ and $B$. 
\end{lemma}

The rest of the section is devoted to proving Lemma \ref{lem:main_theorem} by contradiction, with proofs partially omitted due to space limits. For the rest of the section fix a set $A \cup B$ and let the partition $A,B$ correspond to the optimal solution to the $2$-means problem on $A\cup B$. For sake of contradiction suppose more than $\frac{3}{7}|A|$ points in $A$ are low-revenue points. We will show that such a split $A \cup B \rightarrow(A,B)$ cannot be optimal for the $2$-means objective.  Indeed, we will show that  another split has a smaller $2$-means objective value, proving the lemma.

Say we have $i \in A$ and $j \in B$, such that $rev(i,j)<\frac{1}{10}$. Let $H$ be the hyperplane such that $H=\{y: d(y,\cen(A))=d(y,\cen(B))\}$. Then, $H$ separates the Euclidean space into two half-spaces: $H^+=\{y: d(y,\cen(A))\geq d(y,\cen(B))\}$ and $H^-=\{y: d(y,\cen(A))\leq d(y,\cen(B))\}$. By the assumption that the split $A\cup B \rightarrow (A,B)$ is the optimal $2$-means solution, we have $A \subseteq H^+$, and $B \subseteq H^-$. Next we show the following structural lemma.  This lemma says that if $rev(i,j)$ is small then $d(i,\cen(A))$ and $d(j,\cen(B))$  are within a constant factor of each other, which is close to $1$. Geometrically, this implies that both $i$ and $j$ are located close to the hyperplane $H$. See Figure \ref{fig:optimal_2_means_fig} for an illustration.  The following lemma's proof is in the appendix.

\begin{figure}[H]
\label{fig:optimal_2_means_fig}
\centering 
\includegraphics[height=3.8cm]{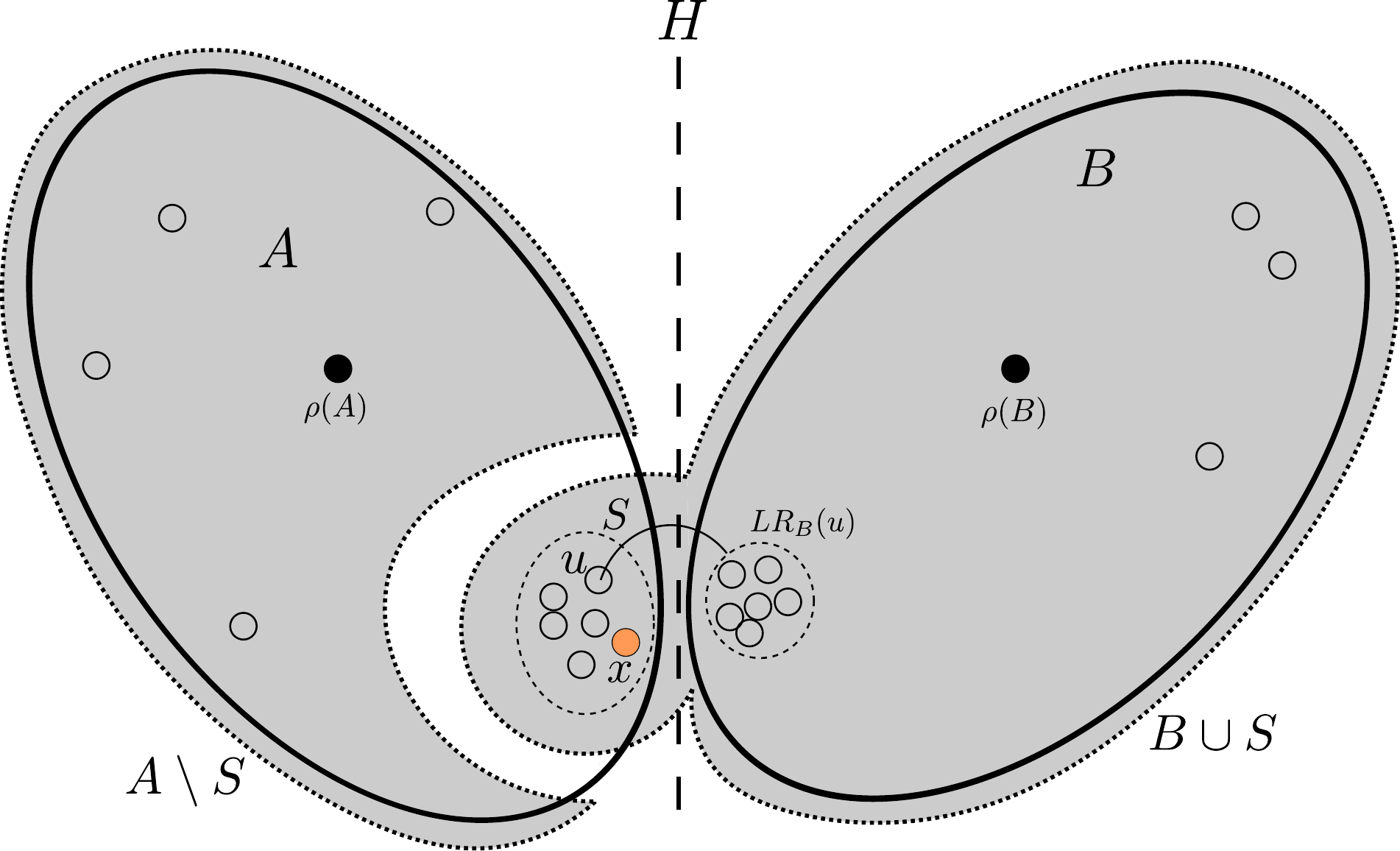} 
\caption{Proof by constructing a better $2$-means solution. The bold dashed line in the middle is the hyperplane $H$. The two bold ellipses are clusters $A$ and $B$ , respectively. The dashed ellipse in $A$ is the set $S$, and the dashed ellipse in $B$ is the low-revenue set $LR_B(u)$ for point $u \in A$ in $B$. $S$ and $LR_B(u)$ are both close to the separating hyperplane $H$. A new partition $A \cup B \rightarrow (A\setminus S, B \cup S)$ is constructed, represented by the two grey areas.}
\end{figure}

\begin{lemma} \label{lem:two_distance_close}
Consider any $i\in A$ and $j\in B$. If $rev(i,j)<\frac{1}{10}$, we have $\frac{9}{10}d(i,\cen(A))< d(j,\cen(B))< \frac{10}{9}d(i,\cen(A))$, and $\frac{9}{10}d(j,\cen(B))< d(i,\cen(A))< \frac{10}{9}d(j,\cen(B))$, and $d(i,j)<\frac{1}{9}\min \{d(i,\cen(A)), d(j,\cen(B)) \}$.
\end{lemma}

Let $S$ be the subset of low-revenue points in $A$. By assumption, $|S|>\frac{3}{7}|A|$. The next lemma establishes that any two points in $S$ are very close to each other as compared to their distance to the centroid $\cen(A)$. The following lemma's proof is in the appendix.
\begin{lemma}\label{lem:small_S}
Let $S$ be the low-revenue points in $A$. For any two points $u,v\in S$, $d(u,v) \leq \frac{2}{9}\max \{d(u,\cen(A)), d(v, \cen(A)) \}$.
\end{lemma}
Let $x$ be the point in $S$ such that $x \in \argmax_{u \in S}d(u,\cen(A))$, the farthest points from $\cen(A)$ in $S$. Notice that, $d(x, \cen(A)) > 0$, otherwise it implies $S$ is overlapping with $\cen(A)$, for any $u \in S$ and $v \in LR_B(u)$, by Lemma \ref{lem:two_distance_close}, we have $d(v, \cen(B))=0$, but this implies $rev(u,v)=1$. Therefore, $d(x, \cen(A))>0$.

Lemma \ref{lem:small_S} implies that $\forall u \in S$, $d(u,x) \leq \frac{2}{9}d(x,\cen(A))$. This result tells us the set $S$ is contained in a ball centered at $x$, with radius $\frac{2}{9}d(x,\cen(A))$. So we can bound the distance between centroid of $S$, $\cen(S)$ and $\cen(A)$ using convexity of the $\ell_2$ norm. The following lemma's proof is omitted due to space.

\begin{lemma} \label{lem:cS_far_from_cA}
Let $S$ be the low-revenue points in $A$ and $x \in \argmax_{u \in S}d(u,\cen(A))$.  It is the case that $d(\cen(S),\cen(A))\geq \frac{7}{9}d(x,\cen(A))$.
\end{lemma}

This is proved by combining Lemma \ref{lem:small_S} with the convexity of $l_2$ norm. Notice that $\cen(A)$ is a convex combination of all points in $A$, Jensen's inequality gives us the conclusion. Since we proved $\cen(S)$ is far from $\cen(A)$. Next, we upper-bound $d(\cen(S), \cen(B))$. Recall that points in the set $S$ are far away from $\cen(A)$, but close to the hyperplane $H=\{y: d(y,\cen(A))=d(y,\cen(B))\}$. The following lemma's proof is in the appendix.
\begin{lemma} \label{lem:S_in_the_middle}
Let $S$ be the low-revenue points in $A$.  For any $u \in S$, $d(u,\cen(B)) \leq \frac{11}{9}d(x,\cen(A))$.
\end{lemma}

Therefore, we can upper bound $d(\cen(S),\cen(B))$: $d(\cen(S),\cen(B)) \leq \frac{\sum_{u \in S}d(u,\cen(B))}{|S|} \leq \frac{11}{9}d(x, \cen(A))$.  The first inequality follows by definition of a centroid.  The second from Lemma~\ref{lem:S_in_the_middle}. This, combined with $d(\cen(S),\cen(A))\geq \frac{7}{9}d(x,\cen(A))$ from Lemma~\ref{lem:cS_far_from_cA}, gives us the following: $\frac{d^2(\cen(S),\cen(A))}{d^2(\cen(S),\cen(B))}\geq (\frac{7}{9})^2/(\frac{11}{9})^2=\frac{49}{121}$. Recall that $\Delta_k(U)$ denotes the  optimal $k$-means value for a set $U$. Let $S_1$ and $S_2$ be two sets. We quote the following lemma from \citet{ostrovsky2012effectiveness}.

\begin{lemma}[\cite{ostrovsky2012effectiveness}]\label{lem:one_means}
For any two sets of points $S_1$ and $S_2$ it is the case that $\Delta_1(S_1 \cup S_2)=\Delta_1(S_1)+\Delta_1(S_2)+\frac{|S_1||S_2|}{|S_1|+|S_2|}d^2(\cen(S_1),\cen(S_2))$.
\end{lemma}

With this in place Lemma \ref{lem:main_theorem} can be shown. In general, we show this by take the set $S$ away from $A$ and assign it into cluster $B$ instead, and prove that this is a better $2$-means solution than the previous one.  Due to space, the proof of Lemma \ref{lem:main_theorem} is omitted.

\section{Randomly Partitioning Poorly Approximates the Revenue Objective}
\label{sec:random}

Consider the following  algorithm  which can create undesirable  trees. The \emph{Random} algorithm splits a set $S$ into $(S_1,S_2)$ by flipping an independent, fair coin for each point in $S$.  If the coin comes up heads then the point gets added to $S_1$, and otherwise gets added to $S_2$. The algorithm is intuitively undesirable  because it does not take the structure of the input into the construction of the solution. Further, the solutions produced do not give much insight into the data.

While intuitively bad, this algorithm is known to be a  $\frac{1}{3}$-approximation for the objective of \citet{DBLP:conf/nips/MoseleyW17}  with similarity scores and it is a $\frac{2}{3}$-approximation for \ckmm objective for dissimilarity scores.  These results hold for any set of similarity or dissimilarity scores, regardless of if they form a metric.   

We show that the our objective does not have this shortcoming. The approximation ratio of the Random algorithm is at most $O(\frac{1}{n^\epsilon})$ for a constant $\epsilon >0$, indicating that it performs very poorly, as is stated by Theorem \ref{thm:random_is_bad}. Proof is omitted.

\begin{theorem} \label{thm:random_is_bad}
Let $OPT(V)$ be the optimal solution for $V$. Let the expected revenue be $\mathbb{E}_T[rev_T(V)]$ for set $V$. Then, there exists a construction of $V$, such that for a constant $\epsilon \in (0,1)$, $\mathbb{E}_T[rev_T(V)] = O(\frac{1}{n^\epsilon})\cdot OPT(V)$. 
\end{theorem}

\section{Objectives for Data in Metric Space}
\label{sec:everytree} 
This section studies data with similarity/dissimilarity scores in a metric space.  First we investigate the \ckmm objective for hierarchical clustering on point sets using dissimilarity scores.  Recall that this objective is the same as the Dasgupta \cite{DBLP:conf/stoc/Dasgupta16} objective except the minimization is swapped for a maximization and the similarity scores are swapped for dissimilarity scores. We also study the Dasgupta objective \cite{DBLP:conf/stoc/Dasgupta16} for similarity scores. We show for each case that if the pairwise similarity/dissimilarity scores form \emph{any metric}, then \emph{every} tree is a at most a factor $2$ from optimal.  

For a tree $T$ let $T[i \lor j]$ denote the subtree rooted at the least common ancestor of $i$ and $j$, and $|\verb+leaves+(T[i \lor j])|$ denote the number of leaves of $T[i \lor j]$.  Recall that the \ckmm objective is the following: $\max_{T} \sum_{i,j \in V}d(i,j)|\verb+leaves+T[i \lor j]|$.

\begin{table*}[ht]
    \centering
    \begin{tabular}{lllll}
    \hline
    Algorithm & $(\Hat{\mu}_1, \Hat{\sigma}_1)$-Census & $(\Hat{\mu}_2, \Hat{\sigma}_2)$-Census & $(\Hat{\mu}_1, \Hat{\sigma}_1)$-Bank & $(\Hat{\mu}_2, \Hat{\sigma}_2)$-Bank \\
    \hline
    bisecting k-means & (4.931e5, 304.980) & (1.094e12, 1.714e11) & (4.912e5, 474.451) & (1.049e12, 1.158e11) \\ 
    average-linkage & (4.900e5, 1.151e3) & (1.093e12, 1.710e11) & (4.907e5, 802.665) & (1.052e12, 1.163e11) \\ 
    single-linkage & (4.869e5, 1.392e3) & (1.094e12, 1.712e11) & (4.818e5, 1.365e3) & (1.035e12, 1.168e11) \\
    Random & (1.311e5, 1.072e4) & (7.463e11, 1.152e11)  & (3.339e5, 8.825e3) & (7.789e11, 7.993e10)\\
    upper bound & (499500, 0) & (1.119e12, 1.725e11) & (499500, 0) & (1.167e12, 1.199e11) \\
    \hline
    \end{tabular}
    \caption{Summary of stats for all algorithms, on Census and Bank}
    \label{tab:exp_data}
\end{table*}

In \citet{cohen2019hierarchical}, it has been proved that any solution is a constant approximation of the optimal solution for \ckmm objective, given that the distance is a metric. Here we prove a stronger conclusion:

\begin{theorem} \label{thm:cohen_addad_bad}
Any solution is a $\frac{1}{2}$-approximation for \ckmm objective if the distance $d(i,j)$ is a metric, i.e., it satisfies triangle inequality.
\end{theorem}

Next consider the objective in Dasgupta~\cite{DBLP:conf/stoc/Dasgupta16}.  Here each pair of points $i$ and $j$ have a similarity score $w_{ij}$ where higher weights mean points are more similar.  Recall from the introduction that Dasgupta's objective is $\min_{T} \sum_{i,j \in V}w_{ij}|\verb+leaves+T[i \lor j]|$.

We show the following corollary that follows from the proof of the prior theroem. 

\begin{corollary} \label{cor:dasgupta_bad_metric}
If the similarity score in the setting of \citet{DBLP:conf/stoc/Dasgupta16} is a metric, any hierarchical clustering tree is a $2$-approximation for the objective in \citet{DBLP:conf/stoc/Dasgupta16}: $\min_{T}cost_T(V)=\sum_{1 \leq i < j \leq n}w_{ij}|\verb+leaves+(T[i \lor j])|$.
\end{corollary}

We note that for similarity scores, it is not a standard assumption that data lies in a metric space.  Thus, this corollary is perhaps interesting to understand the structure of the objective.  However, it does not suggest that any tree will be $2$-approximate for most data sets with similarity scores.

\section{Empirical Results}

The goal of this section is to study the performance of different algorithms for the new objective empirically. The experimental results support the following claims:
\begin{itemize}
    \item  Algorithms that are popular in practice give high revenue for the new objective, with bisecting k-means performing the best.  This demonstrates that the objective value is correlated with algorithms that perform well and highly connected to the bisecting k-means algorithm, as the theory suggests.
    \item Random algorithm, as mentioned in previous section, performs poorly for the new objective.
\end{itemize}

\medskip
\noindent\textbf{Data sets:}
We use two data sets from the UCI data repository: \emph{Census}\footnote{https://archive.ics.uci.edu/ml/datasets/census+income} and \emph{Bank}\footnote{https://archive.ics.uci.edu/ml/datasets/Bank+Marketing}. Only the numerical features are used. 

\medskip
\noindent \textbf{Algorithms studied:}
We study four algorithms\footnote{https://github.com/wangyuyan2333/hier\_clustering\_split\_rev\_ \\ obj\_test.git}: bisecting k-means, average-linkage, single-linkage, and Random. In each experiment, we subsample $2000$ data points from the given data set and run the algorithms with subsampled data. We conduct five experiments with each data set and report the mean and variance. Since optimal $2$-means solution is intractible, in practice we import the k-means implementation from package \emph{Scikit-learn}\footnote{https://scikit-learn.org/stable/modules/generated/sklearn\\.cluster.KMeans.html}, which uses Lloyd's algorithm seeded with k-means++ for each split.

\medskip
\noindent \textbf{Results:}
Table \ref{tab:exp_data} shows the comparison between performance for our objective and the \ckmm objective. For each algorithm, the columns $(\Hat{\mu}_1, \Hat{\sigma}_1)$ and $(\Hat{\mu}_2, \Hat{\sigma}_2)$ denote the mean and standard deviation for our objective and \ckmm objective respectively, calculated over results of the five experiments.

Regarding the new objective, the results show bisecting k-means performs the best of the four algorithms for it. Further, bisecting k-means is within $1\%$ of the upper bound on the optimal solution for the objective, which is the total number of pairs of data points. This suggests that the objective is closely related to bisecting k-means, as the theory suggests. It also shows that experimentally bisecting k-means performs much better than the approximation ratio established.

All the three algorithms which are popular in practice perform well for our objective, with bisecting k-means performing marginally better than average-linkage and single-linkage on average. Moreover, bisecting k-means also has the smallest standard deviation across different subsamples. Random is significantly worse, with potentially over 30 times more loss compared to optimal than the other algorithms. This perhaps suggests that trees created by good algorithms perform well for the objective and poorly constructed trees do not perform well.

Compared with the \ckmm objective from prior work, the results further show that average-linkage performs slightly better than bisecting k-means for the \ckmm objective. This result matches the theory, which suggests this objective is closer to average-linkage than bisecting k-means. Again, all three algorithms used in practice perform well for \ckmm. However, Random also gives about $2/3$ of the upper-bound, as the theoretical bound suggests. This perhaps shows the \ckmm objective gives similar judgements on algorithm performance with our objective, the latter showing a more significant gap between Random and the other three algorithms.

\section{Conclusion}
This paper gives a new objective function for hierarchical clustering designed to mathematically capture the principle used to motivate most divisive algorithms. That is, comparing inter vs. intra cluster distances at splits in the tree. 

The paper proved a close relationship between the objective and the  bisecting $k$-means algorithm.  This was done by showing  the bisecting $k$-means provably optimizes the objective. This helps to understand the structures of trees produced using bisecting $k$-means.  

The results in this paper leave  directions for future work.  How tight can the approximation ratio be for the $k$-means algorithm?  How do other hierarchical clustering algorithms perform for this objective? Can we improve on the bisecting $k$-means algorithm to better optimize the objective?

\bibliographystyle{aaai}
\bibliography{ref}

\begin{thebibliography}{}

\bibitem[\protect\citeauthoryear{Ackerman and
  Ben{-}David}{2016}]{DBLP:journals/jmlr/AckermanB16}
Ackerman, M., and Ben{-}David, S.
\newblock 2016.
\newblock A characterization of linkage-based hierarchical clustering.
\newblock {\em Journal of Machine Learning Research} 17:232:1--232:17.

\bibitem[\protect\citeauthoryear{Awasthi, Blum, and
  Sheffet}{2012}]{AwasthiBS12}
Awasthi, P.; Blum, A.; and Sheffet, O.
\newblock 2012.
\newblock Center-based clustering under perturbation stability.
\newblock {\em Inf. Process. Lett.} 112(1-2):49--54.

\bibitem[\protect\citeauthoryear{Balcan, Blum, and Gupta}{2013}]{BalcanBG13}
Balcan, M.; Blum, A.; and Gupta, A.
\newblock 2013.
\newblock Clustering under approximation stability.
\newblock {\em J. {ACM}} 60(2):8:1--8:34.

\bibitem[\protect\citeauthoryear{Balcan, Blum, and Vempala}{2008}]{BalcanBV08}
Balcan, M.; Blum, A.; and Vempala, S.
\newblock 2008.
\newblock A discriminative framework for clustering via similarity functions.
\newblock In {\em Proceedings of STOC},  671--680.

\bibitem[\protect\citeauthoryear{Carlsson and M{\'{e}}moli}{2010}]{CarlssonM10}
Carlsson, G.~E., and M{\'{e}}moli, F.
\newblock 2010.
\newblock Characterization, stability and convergence of hierarchical
  clustering methods.
\newblock {\em Journal of Machine Learning Research} 11:1425--1470.

\bibitem[\protect\citeauthoryear{Charikar and
  Chatziafratis}{2017}]{CharikarC17}
Charikar, M., and Chatziafratis, V.
\newblock 2017.
\newblock Approximate hierarchical clustering via sparsest cut and spreading
  metrics.
\newblock In {\em Proceedings of SODA},  841--854.

\bibitem[\protect\citeauthoryear{Charikar \bgroup et al\mbox.\egroup
  }{2019}]{pmlr-v89-charikar19a}
Charikar, M.; Chatziafratis, V.; Niazadeh, R.; and Yaroslavtsev, G.
\newblock 2019.
\newblock Hierarchical clustering for euclidean data.
\newblock In Chaudhuri, K., and Sugiyama, M., eds., {\em PMLR}, volume~89,
  2721--2730.
\newblock PMLR.

\bibitem[\protect\citeauthoryear{Charikar, Chatziafratis, and
  Niazadeh}{2019}]{CharikarCN19}
Charikar, M.; Chatziafratis, V.; and Niazadeh, R.
\newblock 2019.
\newblock Hierarchical clustering better than average-linkage.
\newblock In {\em Proceedings of SODA},  2291--2304.

\bibitem[\protect\citeauthoryear{Chatziafratis, Niazadeh, and
  Charikar}{2018}]{ChatziafratisNC18}
Chatziafratis, V.; Niazadeh, R.; and Charikar, M.
\newblock 2018.
\newblock Hierarchical clustering with structural constraints.
\newblock In {\em Proceedings of the ICML},  773--782.

\bibitem[\protect\citeauthoryear{Cohen-addad \bgroup et al\mbox.\egroup
  }{2019}]{cohen2019hierarchical}
Cohen-addad, V.; Kanade, V.; Mallmann-trenn, F.; and Mathieu, C.
\newblock 2019.
\newblock Hierarchical clustering: Objective functions and algorithms.
\newblock {\em J. ACM} 66(4):26:1--26:42.

\bibitem[\protect\citeauthoryear{Dasgupta and Long}{2005}]{DasguptaL05}
Dasgupta, S., and Long, P.~M.
\newblock 2005.
\newblock Performance guarantees for hierarchical clustering.
\newblock {\em J. Comput. Syst. Sci.} 70(4):555--569.

\bibitem[\protect\citeauthoryear{Dasgupta}{2016}]{DBLP:conf/stoc/Dasgupta16}
Dasgupta, S.
\newblock 2016.
\newblock A cost function for similarity-based hierarchical clustering.
\newblock In {\em STOC},  118--127.

\bibitem[\protect\citeauthoryear{Ghoshdastidar, Perrot, and von
  Luxburg}{2018}]{abs-1811-00928}
Ghoshdastidar, D.; Perrot, M.; and von Luxburg, U.
\newblock 2018.
\newblock Foundations of comparison-based hierarchical clustering.
\newblock {\em CoRR} abs/1811.00928.

\bibitem[\protect\citeauthoryear{Moseley and
  Wang}{2017}]{DBLP:conf/nips/MoseleyW17}
Moseley, B., and Wang, J.
\newblock 2017.
\newblock Approximation bounds for hierarchical clustering: Average linkage,
  bisecting k-means, and local search.
\newblock In {\em Advances in Neural Information Processing Systems, 2017,},
  3097--3106.

\bibitem[\protect\citeauthoryear{Murtagh and
  Contreras}{2012}]{DBLP:journals/widm/MurtaghC12}
Murtagh, F., and Contreras, P.
\newblock 2012.
\newblock Algorithms for hierarchical clustering: an overview.
\newblock {\em Wiley Interdisc. Rew.: Data Mining and Knowledge Discovery}
  2(1):86--97.

\bibitem[\protect\citeauthoryear{Murtagh}{1983}]{DBLP:journals/cj/Murtagh83}
Murtagh, F.
\newblock 1983.
\newblock A survey of recent advances in hierarchical clustering algorithms.
\newblock {\em Comput. J.} 26(4):354--359.

\bibitem[\protect\citeauthoryear{Ostrovsky \bgroup et al\mbox.\egroup
  }{2012}]{ostrovsky2012effectiveness}
Ostrovsky, R.; Rabani, Y.; Schulman, L.~J.; and Swamy, C.
\newblock 2012.
\newblock The effectiveness of lloyd-type methods for the k-means problem.
\newblock {\em Journal of the ACM (JACM)} 59(6):28.

\bibitem[\protect\citeauthoryear{Plaxton}{2006}]{Plaxton06}
Plaxton, C.~G.
\newblock 2006.
\newblock Approximation algorithms for hierarchical location problems.
\newblock {\em J. Comput. Syst. Sci.} 72(3):425--443.

\bibitem[\protect\citeauthoryear{Roy and Pokutta}{2017}]{RoyP17}
Roy, A., and Pokutta, S.
\newblock 2017.
\newblock Hierarchical clustering via spreading metrics.
\newblock {\em JMLR} 18:88:1--88:35.

\bibitem[\protect\citeauthoryear{Steinbach \bgroup et al\mbox.\egroup
  }{2000}]{steinbach2000comparison}
Steinbach, M.; Karypis, G.; Kumar, V.; et~al.
\newblock 2000.
\newblock A comparison of document clustering techniques.
\newblock In {\em KDD workshop on text mining}, volume 400,  525--526.
\newblock Boston.

\bibitem[\protect\citeauthoryear{Wang and
  Wang}{2018}]{DBLP:journals/corr/abs-1812-02715}
Wang, D., and Wang, Y.
\newblock 2018.
\newblock An improved cost function for hierarchical cluster trees.
\newblock {\em CoRR} abs/1812.02715.

\bibitem[\protect\citeauthoryear{Zhao, Karypis, and
  Fayyad}{2005}]{DBLP:journals/datamine/ZhaoKF05}
Zhao, Y.; Karypis, G.; and Fayyad, U.~M.
\newblock 2005.
\newblock Hierarchical clustering algorithms for document datasets.
\newblock {\em Data Min. Knowl. Discov.} 10(2):141--168.

\end{thebibliography}

\clearpage
\begin{appendix}
\section*{Appendix}
\section{Ground-truth Inputs}
\label{sec:proving_ground_truth}
This section proves the \probname function works for ground-truth inputs proposed in \citet{cohen2019hierarchical}.

\begin{proofof}[Lemma~\ref{lem:generating_tree_cuts_biggest_distances}]
The if direction is true, since from top to bottom, at each split the tree $T$ cuts only  the longest distances in the current set of points, both properties in Definition~\ref{def:generating_tree} trivially hold. We prove the only if direction. For every $i \in A$ and $j \in B$, $LCA_T(i, j)$ is always the node representing $A \cup B$, so $d(i,j) = W(LCA_T(i, j))$ is always the same value. To show it is the maximum distance in all the pairwise distances in $A \cup B$, assume that is not the case. Then some pair of vertices of maximum distance is contained in the subgraph induced by $A$ or $B$, which means it will be cut in the subtree rooted at $A$ or $B$, say it is cut at root $N_2$, and let $N_1 = LCA_T(i, j)$. $N_1$ is on the way from $N_2$ to the path, but $W(N_1) < W(N_2)$, contradicting property (1).
\end{proofof}

\begin{proofof}[Theorem~\ref{thm:always_generating_tree}]
We prove the theorem by constructing one such tree in the following way. Say given a set $S$, we separate it into two sets $L$ and $R$. 
\begin{enumerate}
    \item Pick a pair of points $(i,j)$ with longest distance. Put $i$ into $L$ and $j$ into $R$.
    \item For any point $x \in S$, either $d(i,x) < d(i,j)$ or $d(i,x) = d(i,j)$ since $d(i,j)$ is chosen to be the maximum. If $d(i,x) = d(i,j)$ put $x$ into $R$, otherwise put it into $L$.
\end{enumerate}
To argue that all points in $L$ and $R$ are of distance $d(i,j)$ from each other, notice that if $L$ only contains $x$ the theorem trivially holds. Otherwise,  apparently by construction we also have $\forall{y \in R}, d(i,y) = d(i,j)$. Now take any two points $x \in L, y \in R$, we further prove $d(x,y) = d(i,j)$. Observe that $d(i,x) < d(i,j)$ but $d(i,y) = d(i,j)$. Again by definition of ultrametric, in the triangle formed by $i,x,y$, we have $d(x,y) = d(i,y) > d(i,x)$. By Lemma~\ref{lem:generating_tree_cuts_biggest_distances}, this is a generating tree for $G$.
\end{proofof}

\begin{proofof}[Theorem \ref{thm:generating_tree_optimal}]

Given any split in the tree $A \cup B \rightarrow (A,B)$, for any $i \in A$ and $j \in B$, we prove that $d(i,\cen(A)) \leq d(i,j)$ and $d(j,\cen(B)) \leq d(i,j)$. As a result, $rev(i,j) = \frac{d(i,j)}{\max \{d(i,\cen(A)), d(j, \cen(B)), d(i,j)\}} = 1$.

Let's focus on $A$ for the time being. By Lemma~\ref{lem:generating_tree_cuts_biggest_distances}, $\forall{x \in A}, d(i,x) \leq d(i,j)$. By convexity of norms, $d(i,\cen(A)) = d(i,\frac{\sum_{x \in A}x}{|A|}) \leq \frac{\sum_{x \in A}d(x,i)}{|A|} \leq d(i,j)$. The other inequality, $d(j, \cen(A)) \leq d(i,j)$, can be proved in the same way.
\end{proofof}

\section{Proving Bisecting $k$-means Optimizes the Revenue Objective}
\label{sec:sup_2_means}
This section covers the omitted proofs in Section \ref{sec:2_means_approx}.

\begin{proofof}[Lemma \ref{lem:two_distance_close}]
Say that $rev(i,j)<\frac{1}{10}$. Without loss of generality assume that $d(i,\cen(A)) \geq d(j,\cen(B))$. This and the definition of revenue give  $d(i,j) < \frac{1}{10}d(i,\cen(A))$. Since $A$ and $B$ is the optimal $2$-means partition, $d(i,\cen(A)) \leq d(i,\cen(B))$ and $d(j,\cen(B)) \leq d(j,\cen(A))$. The triangle inequality gives,
\begin{align*}
d(j,\cen(B))&\geq d(i,\cen(B))-d(i,j) \geq d(i,\cen(A)) - d(i,j)\\
&> d(i,\cen(A))-\frac{1}{10}d(i,\cen(A))=\frac{9}{10}d(i,\cen(A))
\end{align*}
An analogous proofs shows  $d(i,\cen(A))>\frac{9}{10}d(j,\cen(B))$. The last inequality in the lemma follows immediately from these two inequalities.
\end{proofof}

\begin{proofof}[Lemma \ref{lem:small_S}]
Recall that $LR_B(u)$  is the set of points in $w \in B$ such that  $rev(u,w) < \frac{1}{10}$. Similarly for $LR_B(v)$. Knowing that $|LR_B(u)| > \frac{1}{2}|B|$ and $|LR_B(v)| > \frac{1}{2}|B|$, there exists some point $w\in B$, such that $rev(u,w) < \frac{1}{10}$ and $rev(v,w)< \frac{1}{10}$. Without loss of generality suppose $d(u,\cen(A)) \geq d(v, \cen(A))$. We want to show $d(u,v) \leq \frac{2}{9}d(u,\cen(A))$, notice that $d(u,v) \leq d(u,w)+d(v,w)$, and we have $d(u,w) \leq \frac{1}{9}d(u,\cen(A))$ and $d(v,w) \leq \frac{1}{9}d(v,\cen(A))$, respectively, by Lemma \ref{lem:two_distance_close}. Note that $d(v,\cen(A))\leq d(u,\cen(A))$, so $d(v,w) \leq \frac{1}{9}d(u,\cen(A))$, and we conclude that $d(u,v)\leq \frac{2}{9}d(u,\cen(A))$.
\end{proofof}

\begin{proofof}[Lemma \ref{lem:cS_far_from_cA}]
\begin{align*}
d(\cen(S),x) &= d(\frac{\sum_{u \in S}u}{|S|}, x) \leq \frac{\sum_{u \in S}d(u,x)}{|S|}\\
& \leq \frac{2}{9}d(x,\cen(A))
\end{align*}
As a result the triangle inequality gives, $d(\cen(S),\cen(A))\geq d(x,\cen(A))-d(x,\cen(S))\geq \frac{7}{9}d(x,\cen(A))$.
\end{proofof}

\begin{proofof}[Lemma \ref{lem:S_in_the_middle}]
 Since $u \in S$, there exists $w \in B$, s.t. $d(u,w) < \frac{1}{10}\max \{d(u,\cen(A)), d(w,\cen(B)) \}$. By triangle inequality, we have $d(u,\cen(B))\leq d(w,\cen(B))+d(u,w)$. Since $rev(u,w)<\frac{1}{10}$, by Lemma \ref{lem:two_distance_close}, $d(w,\cen(B)) \leq \frac{10}{9}d(u,\cen(A))$ and $d(u,w) \leq \frac{1}{9}d(u,\cen(A))$. Therefore, $d(u,\cen(B))\leq \frac{11}{9}d(u,\cen(A)) \leq \frac{11}{9}d(x,\cen(A))$.
\end{proofof}

\section{Proving Random is Bad}
\label{sec:sup_random_bad}
This section is devoted to proving Theorem \ref{thm:random_is_bad}. 

\medskip
\noindent \textbf{Constructing the input Point Set:} The input consists of two \emph{unbalanced} sets of points $A$ and $B$ where $|A|=n^2$, $|B|=n$.  We assume that the points in $A$ and $B$ are very far away but the intra-cluster distance is small.  We will set this parameter later.   For simplicity $A$ consists of points all in the same location and the same for $B$. Let $V=A \cup B$ be the entire point set.



\subsection{An Upper Bound on the Performance of Random}

Before we argue Random is bad, we give the definition of ``clean split''. Intuitively, a split should be considered clean if it doesn't separate points close to each other when there are far away pairs.
\begin{definition}
We define a split $S \rightarrow (S_1, S_2)$ to be \emph{clean} if it satisfies one of the following conditions:
\begin{enumerate}
    \item If $S \subseteq A$ or $S \subseteq B$.
    \item If $S_1 \subseteq A, S_2 \subseteq B$, or $S_1 \subseteq B, S_2 \subseteq A$.
\end{enumerate}
\end{definition}

Based on the result that every tree is gaining full revenue for an ultrametric from Section~\ref{sec:proving_ground_truth}, it is easy to see that optimal tree can get a revenue of $OPT(V):= \frac{(n^2+n)(n^2+n-1)}{2}=\Theta(n^4)$ for the whole point set. The optimal tree splits $A$ from $B$ in the root split, and then can do anything on the remaining portion of the tree.

Before formally prove this theorem we make some quick observations. First, we don't need to care about the pairs $(i,j)$ where $i \in A$ and $j \in B$ because the number of such pairs is $\Theta(n^3)$, even if we gain full revenue for them, it doesn't affect the approximation ratio. For the same reason we don't care about points $(i,j)$ such that $i,j \in B$. So, we only need to discuss how much revenue we can get from separating all the pairs inside $A$ in expectation for Random.

With this in mind, we will use Chernoff bounds to argue that for $\Theta(\log{n})$ rounds, Random splits each node in half with high probability,  which causes us to lose a lot of revenue.

\begin{lemma} \label{lem:equal_split}
Suppose we have a set $S$ with $m$ points, and use Random to split it into $S_1$ and $S_2$. Then, for $i=1,2$
$$\mathbb{P}(||S_i|-\frac{m}{2}| \leq \sqrt{m \log{m}}) \geq 1-\frac{2}{m^2}$$
\end{lemma}

\begin{proofof}[Lemma \ref{lem:equal_split}]
Consider $m$ i.i.d. Rademacher variables $X_j$. Then from Chernoff's bound, we know that
$$\mathbb{P}(|\sum_{j=1}^m X_j| \geq t) \leq 2 \exp(-\frac{t^2}{2m})$$
Random is treating each point $j$ as a Rademacher variable by assigning 
$$X_j=
\begin{cases}
+1 & \text{if } j \text{ is assigned to } S_1\\
-1 & \text{if } j \text{ is assigned to } S_2
\end{cases}$$
Then, for $i=1,2$,
\begin{align*}
    \mathbb{P}(||S_i|-\frac{m}{2}| \geq \sqrt{m \log(m)}) & = \mathbb{P}(|\sum_{i=1}^m X_i| \geq 2\sqrt{m \log(m)})\\
    & \leq 2\exp(2 \log(m))=\frac{2}{m^2}
\end{align*}
\end{proofof}

Next, we define ``almost-equal" splits, which refers splits such that the points from $A$ and $B$ in the parent node is almost split equally in its two children.
\begin{definition} \label{def:almost_equal}
Given a set $S$, let $S^A$ and $S^B$ denote the points from $A$ and $B$ in $S$, respectively. If a split $S \rightarrow (S_1,S_2)$ satisfies the property in Lemma \ref{lem:equal_split}, i.e., for $i=1,2$, let $S_i^A$ and $S_i^B$ denote the set of points from $A$ and $B$ in set $S_i$ respectively, we say this split is \emph{almost equal} if for $i=1,2$:
\begin{enumerate}
    \item $\mathbb{P}(||S_i^A|-\frac{1}{2}|S^A||\leq \sqrt{S^A\log{S^A}})$
    \item $\mathbb{P}(||S_i^B|-\frac{1}{2}|S^B||\leq \sqrt{S^B\log{S^B}})$
\end{enumerate}
Also, for a hierarchical clustering tree, if all the nodes in the first $i$ layers are almost equally split, we call this tree \emph{i-almost equally split}.
\end{definition}
The next lemma bounds the number of points in both $A$ and $B$ in an internal node in $i^{th}$ layer if every split is almost equal for both in the first $i$ layers in the tree, where $i \leq \frac{\log(n)}{2}$.
\begin{lemma} \label{lem:bounds_on_number}
Let $S_i$ be a node in the $i^{th}$ layer of the tree ($i \leq \log{n}/2$). If all the ancestors of $S_i$ is almost-equally split, let $S_i^A$ be the points in $S_i$ in $A$, and $S_i^B$ be the points in $S_i$ in $B$. Then we have $|S_i^A|=\Theta(n^2/2^i)$, $|S_i^B|=\Theta(n/2^i)$. 
\end{lemma}

\begin{proofof}[Lemma \ref{lem:bounds_on_number}]
By induction, we prove a stronger conclusion:
$$\frac{n^2}{2^i}-8\sqrt{\frac{n^2}{2^i} \log(\frac{n^2}{2^i})} \leq |S_i^A| \leq \frac{n^2}{2^i}+8\sqrt{\frac{n^2}{2^i} \log(\frac{n^2}{2^i})}$$
and
$$\frac{n}{2^i}-8\sqrt{\frac{n}{2^i} \log(\frac{n}{2^i})} \leq |S_i^B| \leq \frac{n}{2^i}+8\sqrt{\frac{n}{2^i} \log(\frac{n}{2^i})}$$
We just prove the first claim and the other can be proved in the same way. By induction,
\begin{align*}
    |S_i^A| &\geq \frac{|S_{i-1}^A|}{2}-\sqrt{|S_{i-1}^A| \log (|S_{i-1}^A|)}\\
    &\geq \frac{1}{2}\cdot (\frac{n^2}{2^{i-1}}-8\sqrt{\frac{n^2}{2^{i-1}} \log(\frac{n^2}{2^{i-1}})})\\
    & \qquad -\sqrt{\frac{n^2}{2^{i-1}} \log(\frac{n^2}{2^{i-1}})}\\
    &=\frac{n^2}{2^i}-5\sqrt{2} \cdot \sqrt{\frac{n^2}{2^{i}} (\log(\frac{n^2}{2^{i}})+\log(2))}\\
    &\geq \frac{n^2}{2^i}-8\sqrt{\frac{n^2}{2^i} \log(\frac{n^2}{2^i})}
\end{align*}
And the other side of the inequality can be bounded in the same way.
\end{proofof}

If the condition in Lemma \ref{lem:bounds_on_number} holds, this result tells us that every node in the first $\frac{\log(n)}{2}$ layers is not clean. In other words, for all the pairs of points in $A$ which are separated during the first $\frac{\log(n)}{2}$ layers of the tree, we don't get any revenue. Thus we can upper bound the revenue for points in $A$:
\begin{lemma} \label{lem:bound_if_equal_split}
If the tree $T$ is $\frac{\log(n)}{2}$-almost-equally-split tree, for all the pairs in $A$ the revenue is  $O(n^{4-\epsilon})$ for $\epsilon=\frac{\log(2)}{2}$.
\end{lemma}

We have already proved that if many of the top layers have almost equally split internal nodes, the HC tree has small total revenue. To formally prove Theorem \ref{thm:random_is_bad}, we only need to show that this happens with high probability. Notice that the probability of the tree being not $\frac{\log(n)}{2}$-almost equally split can be bounded by union bounds on the probability of an almost equal split does not happen in any of the first $\frac{\log(n)}{2}$ layers, which is $O(\frac{1}{n^{\epsilon'}})$, where $\epsilon'=2-\frac{3\log{2}}{2}$. This is very low probability, putting everything together, we have Lemma \ref{lem:bound_if_equal_split}.

\begin{proofof}[Lemma \ref{lem:bound_if_equal_split}]
For each internal node in the $(\frac{\log(n)}{2})^{th}$ layer here, the number of points in $A$ is $\Theta(\frac{n^2}{2^{\log(n)/2}})=\Theta(n^{2-\epsilon})$, where $\epsilon = \frac{\log{2}}{2}$, and there are $\Theta(n^{\epsilon})$ such nodes. So, the revenue is bounded by $O(n^{4-\epsilon})$.
\end{proofof}

\begin{proofof}[Theorem \ref{thm:random_is_bad}]
 By Lemma \ref{lem:bound_if_equal_split}, $$\mathbb{E}_T(rev_T(V)|\textit{T is $\frac{\log(n)}{2}$-almost equally split}) = O(n^{4-\epsilon})$$
Then, we only need to lower bound the probability that the tree $T$ is $\frac{\log(m)}{2}$-almost equally split. We show next that this happens with very high probability. Again let $S_i$ denote some node in the $i^{th}$ layer of $T$.
\begin{align*}
&\mathbb{P}(\textit{$S_i$ isn't almost equal split}|
\textit{T is $(i-1)$-almost equally split}) \\
&\qquad \leq \frac{2}{|S_i^B|}+\frac{2}{|S_i^A|} \leq \Theta( \frac{2^i}{n})
\end{align*}
In the $i^{th}$ layer, we have $2^i$ nodes. So we bound the probability of having a tree that's almost equal split in the first $\frac{\log(n)}{2}$ layers as follows:
\begin{align*}
    &\quad \mathbb{P}(\textit{T is $i$-almost equally split})\\
    &=\Pi_{i=1}^{\frac{\log(n)}{2}} \Pi_{S_i\textit{ in the $i^{th}$ layer}} \mathbb{P}(\textit{$S_i$ is almost equally split}|\\
    &\qquad \textit{T is $(i-1)$-almost equally split})\\
    & >\Pi_{i=1}^{\frac{\log(n)}{2}} \Pi_{S_i\textit{ in the $i^{th}$ layer}} (1-\Theta( \frac{2^i}{n}))\\
    & > 1 - \Theta(\sum_{i=1}^{\frac{\log(n)}{2}} \sum_{S_i \textit{ in the $i^{th}$ layer}}  (\frac{2^i}{n})^2)\\
    & = 1- \Theta(\sum_{i=1}^{\frac{\log(n)}{2}} 2^i \cdot \frac{4^i}{n^2})\\
    & = 1-\Theta( \frac{(1+8+8^2+...+8^{\frac{\log(n)}{2}})}{n^2})\\
    &=1-\Theta(2^{\frac{3\log(n)}{2}}/n^2) = 1 - \Theta(\frac{1}{n^{\epsilon'}})
\end{align*}
Where $\epsilon'=2-\frac{3\log{2}}{2}>\epsilon$. So we have $O(\frac{1}{n^{\epsilon'}})$ probability that $T$ is not $i$-almost equally split, in which case the revenue is bounded by $\Theta(n^4)$. 

Therefore, the expectation is bounded by: 
\begin{align*}
    &\qquad \mathbb{E}_T[rev_T(V)]\\
    &\leq \mathbb{P}(T\textit{ is }\frac{\log(n)}{2}\textit{-almost equally split})\Theta(n^{4-\epsilon})\\
    &\qquad +\mathbb{P}(T\textit{ is not  }\frac{\log(n)}{2}\textit{-almost equally split})\Theta(n^4)\\
    &\leq \Theta(n^{4-\epsilon})+\Theta(n^{4-\epsilon'})\\
    &\leq \Theta(n^{4-\epsilon})
\end{align*}
where $\epsilon = \frac{\log(2)}{2}$.
\end{proofof}

\section{Proofs for Cohen-Addad et al. and Dasgupta objectives}
Fix a tree $T$.  Let $LCA(i,j)$ be the least common ancestor of $i$ and $j$ in $T$.  Let $\mathbf{1}\{i,j|k\}$ be an indicator variable indicating whether in the tree $T$ the $LCA(i,j)$ is a descendant of $LCA(i,j,k)$: $\mathbf{1}\{i,j|k\} =1$  if such a relationship holds, and $\mathbf{1}\{i,j|k\}=0$ if otherwise. Equivalently, if $\mathbf{1}\{i,j|k\}=1$, it means during the tree construction, tracing down from the root, $k$ is separated from $i$ and $j$ first, and $i,j$ are separated in a split closer to the bottom of the tree. If the tree is binary, we have the following equality: 
$$\mathbf{1}\{i,j|k\} + \mathbf{1}\{i,k|j\} + \mathbf{1}\{j,k|i\}=1$$
That is, one and only one of the relationships represented by the three indicator variables can hold.

Prior to proving Theorem \ref{thm:cohen_addad_bad}, we cite this result from \citet{DBLP:journals/corr/abs-1812-02715}, which shows the revenue in Cohen-Addad et al. objective can be decomposed onto every triangle:
\begin{lemma} [\cite{DBLP:journals/corr/abs-1812-02715}]\label{lem:triangle_decomposition}
When $|V| \geq 3$,
\begin{align*}
R_T(V)&=\sum_{i,j \in V}d(i,j) |\mathtt{leaves}T[i \lor j]| \\
&=\sum_{\{i,j,k\}\subseteq V}triR_T(i,j,k)+2\sum_{\{p,q\} \subseteq V}d(p,q)
\end{align*}

where $triR_T(i,j,k)$ denotes the revenue on triangle $i,j,k$, defined as follows:
\begin{align*}
    triR_T(i,j,k)=
\begin{cases}
d(i,k)+d(j,k)& \text{if }\mathbf{1} \{i,j|k\}=1\\
d(i,j)+d(j,k)& \text{if } \mathbf{1}\{i,k|j\}=1\\
d(i,j)+d(i,k)& \text{if } \mathbf{1}\{j,k|i\}=1
\end{cases}
\end{align*}
\end{lemma}
By triangle inequality, for each triangle $i,j,k$, we always have $triR_T(i,j,k)\geq \frac{1}{2}(d(i,k)+d(j,k)+d(i,j))$, which will give us Theorem \ref{thm:cohen_addad_bad}.

\begin{proofof}[Theorem \ref{thm:cohen_addad_bad}]
Let $OPT(V)$ denote the optimal value of Cohen-Addad et al. objective for $V$. We have $OPT(V) \leq \sum_{\{p,q\} \subseteq V}|V| \cdot d(p,q)$.

By triangle inequality, it is easy to see that regardless of which of the three relationship holds, we always have $triR_T(i,j,k)\geq \frac{1}{2}(d(i,j)+d(i,k)+d(j,k))$ for any triplet $\{i,j,k\}$. Then, for any $T$,
\begin{align*}
R_T(V)&=\sum_{\{i,j,k\}\subseteq V}triR_T(i,j,k)+2\sum_{\{p,q\} \subseteq V}d(p,q)\\
&\geq \sum_{\{i,j,k\}\subseteq V} \frac{1}{2}(d(i,j)+d(i,k)+d(j,k))\\
& \qquad + 2\sum_{\{p,q\} \subseteq V}d(p,q)\\
&= \frac{1}{2}\sum_{\{p,q\} \subseteq V}d(p,q)(|V|-2) + 2\sum_{\{p,q\} \subseteq V}d(p,q)\\
&>\frac{1}{2}\sum_{\{p,q\} \subseteq V}|V| \cdot d(p,q) \geq \frac{1}{2}OPT(V)
\end{align*}
\end{proofof}

The proof easily gives Corollary~\ref{cor:dasgupta_bad_metric}.

\begin{proofof}[Corollary \ref{cor:dasgupta_bad_metric}]
Similar to Lemma \ref{lem:triangle_decomposition}, we can decompose $cost_T(V)$ in the following way:
\begin{align*}
cost_T(V)&=\sum_{i,j \in V}w_{ij}|\texttt{leaves}T[i \lor j]|\\
&=\sum_{\{i,j,k\}\subseteq V}triC_T(i,j,k)+2\sum_{\{p,q\} \subseteq V}w_{pq}
\end{align*}
where $triC_T(i,j,k)$ is defined as:
$$triC_T(i,j,k)=\begin{cases}
w_{ik}+w_{jk}& \text{if } \mathbf{1}\{i,j|k\}=1\\
w_{ij}+w_{jk}& \text{if } \mathbf{1}\{i,k|j\}=1\\
w_{ij}+w_{ik}& \text{if } \mathbf{1}\{j,k|i\}=1
\end{cases}$$
Then, by triangle inequality, $triC_T(i,j,k) \geq \frac{1}{2}(w_{ij}+w_{ik}+w_{jk})$, so for any given tree $T$,
\begin{align*}
    \min_{T'}cost_{T'}(V) &\geq (\sum_{\{i,j,k\}\subseteq V}\frac{1}{2}(w_{ij}+w_{ik}+w_{jk}))\\
    &\qquad + 2\sum_{\{p,q\}\subseteq V}w_{pq}\\
    &> \frac{1}{2}\sum_{\{p,q\}\subseteq V}w_{pq}(|V|)\\
    &\geq \frac{1}{2} cost_T(V)
\end{align*}
\end{proofof}

\end{appendix}

\end{document}